\documentclass{article}

\usepackage{PRIMEarxiv}

\usepackage[utf8]{inputenc} % allow utf-8 input
\usepackage[T1]{fontenc}    % use 8-bit T1 fonts
\usepackage{hyperref}       % hyperlinks
\usepackage{url}            % simple URL typesetting
\usepackage{booktabs}       % professional-quality tables
\usepackage{amsfonts}       % blackboard math symbols
\usepackage{nicefrac}       % compact symbols for 1/2, etc.
\usepackage{microtype}      % microtypography
\usepackage{lipsum}
\usepackage{fancyhdr}       % header
\usepackage{graphicx}       % graphics
\graphicspath{{media/}}     % organize your images and other figures under media/ folder

\usepackage[utf8]{inputenc}
\usepackage{url}
\usepackage{multirow, makecell}
\usepackage{booktabs}
\usepackage{longtable}
\usepackage{setspace}
\usepackage[T1]{fontenc}
\usepackage{environ}
\usepackage[linesnumbered,ruled,vlined]{algorithm2e}

\usepackage{algpseudocode}
\usepackage{mathtools}
\usepackage{cuted}

\title{Entropy optimized semi-supervised decomposed vector-quantized variational autoencoder model based on transfer learning for multiclass text classification and generation
}

\author{
  Shivani Malhotra, Vinay Kumar \\
  Department of Electronics and Communication Engineering India   \\
  Thapar Institute of Engineering and Technology \\
  Patiala\\
  \texttt{\{smalhotra\_phd18@thapar.edu, vinay.kumar@thapar.edu\}} \\
   \And
  Alpana Agarwal \\
  Department of Electronics and Communication Engineering India  \\
  Thapar Institute of Engineering and Technology  \\
  Patiala\\
  \texttt{alpana@thapar.edu} \\
}

\begin{document}
\maketitle

\begin{abstract}
Semisupervised text classification has become a major focus of research over the past few years. Hitherto, most of the research has been based on supervised learning, but its main drawback is the unavailability of labeled data samples in practical applications. It is still a key challenge to train the deep generative models and learn comprehensive representations without supervision. Even though continuous latent variables are employed primarily in deep latent variable models, discrete latent variables, with their enhanced understandability and better compressed representations, are effectively used by researchers. In this paper, we propose a semisupervised discrete latent variable model for multi-class text classification and text generation. The proposed model employs the concept of transfer learning for training a quantized transformer model, which is able to learn competently using fewer labeled instances. The model applies decomposed vector quantization technique to overcome problems like posterior collapse and index collapse. Shannon entropy is used for the decomposed sub-encoders, on which a variable DropConnect is applied, to retain maximum information. Moreover, gradients of the Loss function are adaptively modified during backpropagation from decoder to encoder to enhance the performance of the model. Three conventional datasets of diversified range have been used for validating the proposed model on a variable number of labeled instances. Experimental results indicate that the proposed model has surpassed the state-of-the-art models remarkably.
%\lipsum[1]
\end{abstract}

% keywords can be removed

\section{Introduction}
The research in the field of probabilistic modeling of various phenomena of data has been an active field of interest in machine learning. Such phenomena can be understood and mathematically described using the probabilistic models \cite{kingmaVAE}.  Moreover, these models prove effective for prediction of unknown variables and several assisted or automated decision making tasks.

Over the past few years, deep latent variable models have become a major focus of research in natural language processing. These models have found applications for learning latent representations of text for supervised learning tasks \cite{yang,Gururangan}, generating text \cite{bowman}, etc. However, it is difficult to perform a well organised inference and learning in these models when the latent variables have intractable posterior distributions. A significant challenge due to these intractable posterior distributions is the indifferentiation and optimization of marginal probability, as determined in fully observed models.

The architecture of Variational Autoencoders (VAEs) presents a computationally productive approach to transform these intractable distributions into tractable forms by introducing a parametric inference model using stochastic gradient descent \cite{kingma}. It is worth noting that during this transformation process, a subset of latent variables is optimally ignored for most pragmatic instances of VAE when it has a sufficiently powerful decoder \cite{chen}. 

Most of the work in deep generative models has modeled continuous latent variables, i.e. as vectors in $\mathbb{R}^D$, partly because of the clarity to accomplish variational inference in VAEs. Nevertheless, these models with discrete latent variables are interesting, considering the fact that discrete latent variables are potentially understandable and generate more compressed representations.

Unfortunately, training of discrete latent variables is challenging as transferring gradient information across discrete units is an onerous task. Although, numerous discretization techniques have been introduced, the present paper focuses on the Vector Quantization (VQ) technique in VAE; i.e. VQ-VAE. This technique helps to evade the posterior collapse problem. Instead of regularizing the latent distribution, VQ-VAE provides a latent representation based on a finite number of centroids. Hence, the capability of the latent representation can be controlled by the number of used centroids, which guarantees that a certain amount of information is preserved in the latent space. It is also emphasized that when the size of discrete latent space is substantial, only a few embedding vectors get trained since the posterior distribution is supported only on a small subset of discrete latent space. As a result, the discrete VAE only uses a small section of latents to compute decoder probability distribution (p(x|z)) and rest of the space is insignificant. This issue is known as index collapse \cite{kaiser}. The architecture proposed in this paper helps to solve this issue to a great extent by decomposing VQ-VAE into $n$ number of varients, thereby utilizing the embedding vectors in a more efficient way.

The key contributions of this paper are as follows:

\begin{enumerate}
    \item We propose a variational autoencoder model based on decomposed vector quantization technique. The proposed model uses a Transformer based encoder and solves the problems of posterior collapse and index collapse.
    \item The concept of Shannon entropy along with the DropConnect technique has been applied on decomposed encoders during training in forward propagation. The gradients of Loss function have been adaptively modified by a factor of 'VQ Loss' during backpropagation so as to preserve maximum information.  
    \item Three experimental datasets have been used to investigate the empirical performance of the proposed model. The experimental results indicate that the proposed model has performed remarkably well relative to other state-of-the-art techniques for variable number of labeled instances. Eventually, some other significant results have been discussed before drawing conclusions for future work.
\end{enumerate}

The framework of the present paper is as follows: Section 2 briefly discusses some of the index terms used in the research work; Section 3 lists out the datasets used along with their elaborative study; Section 4 discusses the existing research work related to the paper; Section 5 introduces and discusses in detail the proposed model along with an algorithm explaining its working ; Section 6 shows the experimental details of our proposed model; Section 7 includes the experimental results and their comparison with other state-of-the-art models; Section 8 discusses the results calculated from the functional blocks of the proposed model; Section 9 concludes the findings and presents some future work proposals.

\section{KEY-WORDS}

\subsection{Variational Autoencoder}

The concept of generative modeling can be considered as an auxiliary task. For instance, future predictions may potentially help to build the useful abstractions of the world that can be employed to predict downstream tasks. The aforementioned research for disentangled, semantically significant and causal factors responsible for variations of data is entitled as unsupervised learning, the purpose for which VAEs have been extensively used. They approximate the Probability Density Function (PDF) of training data and can be regarded as a combination of two independently parameterized models: recognition model or encoder, and the generative model or decoder.

\begin{figure}
%\begin{minipage}[t]{0.50\linewidth}
\centering
\includegraphics[width=0.6\linewidth,height=0.05\textheight]{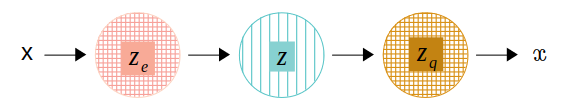}

(a) Standard Autoencoder
%\end{minipage}%
\hfill \hfill
%\begin{minipage}[t]{0.70\linewidth}
\centering
\includegraphics[width= 0.8\linewidth,height=0.10\textheight]{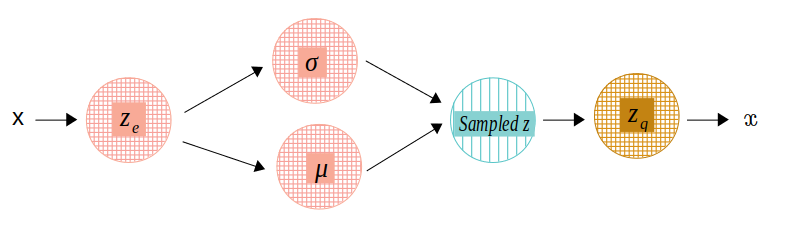}

(b) Variational Autoencoder
%\end{minipage}
\caption{Illustration of difference between a Standard Autoencoder and Variational Autoencoder using probability distribution parameters "$\mu$" and "$\sigma$"}
\label{AEvsVAE}
\end{figure}

According to the Bayes rule, encoder is an approximate inverse of the decoder \cite{kingmaVAE}. The encoder provides an approximation to its posterior over latent random variables to the decoder and updates its parameters inside the repetitive process of expectation maximization learning whereas the decoder enables learning of significant representations of data. In accordance with the conventional autoencoder, the VAE enforces some limitations in the encoding process where the latent vector follows a standard normal distribution and their latent spaces allow easy random sampling and interpolation, as shown in Figure ~\ref{AEvsVAE}. Also, the recognition model is a stochastic function of input variables. In this way, a random latent vector can be generated from a standard normal distribution. 

The architecture of VAE \cite{kingma,rezende} conjointly learns deep latent variable models and analogous inference models using stochastic gradient descent. In order to transform the DLVM’s intractable posterior inference problems into tractable ones, we introduce a parametric inference model, also known as a recognition model. The recognition model parameterises a posterior distribution $q(z|x)$ of discrete latent variable ‘$z$’ for input data ‘$x$’. Apart from this, VAE consists of a generative model with a distribution $p(x|z)$ over input data .

To introduce the recognition model for any kind of variational parameters and improve the variational lower bound on the marginal likelihood for input data, we have:

\begin{equation}
 \resizebox{0.45\textwidth}{!}{$ \log p_\theta(x)= E_{q_\phi(z|x)} \left[\log \left[\frac{p_\theta(x,z)}{q_\phi (z|x)}\right]\right] + E_{q_\phi(z|x)} \left[\log \left[\frac{q_\phi (z|x)}{p_\theta (z|x)}\right]\right ]$ }
\end{equation}

The first term is Equ.(1) is known as ELBO, also written as:
\begin{equation}
 L_{\theta,\phi}(x) = E_{q_\phi(z|x)}[\log p_\theta(x,z) - \log q_\phi(z|x) ] 
\end{equation}

The second term in Equ.(1) is called KL divergence, also represented as: 
\begin{equation}
D_{KL}(q_\phi (z|x))|| p_\theta(z|x) \geq 0 
\end{equation}

It should be observed that the variational inference of our downstream task 'text classification' has two distinct objective functions for unlabeled and labeled data. 
 
\subsection{Discretization Techniques}

Modeling discrete representations is straightforward considering each class has a single value. On the contrary, in continuous latent spaces, it is difficult to normalize the density function and learn the inter-variable dependencies. We know that several significant physical world objects are inherently discrete and Markov switching models (autoregressive models) constituting sequences based on discrete symbols provide remarkable results \cite{oordVQ}. It was shown by \cite{metz} that in spite of the fact that action spaces in reinforcement learning are generally continuous, we can discretize them and use autoregressive models to obtain impressive results. Progressively, this field has presented a growing attention in the field of discrete latent variable models. Recently, some techniques have been introduced which effectively use discrete latent variables: the Gumbel-Max \cite{gumbelmax}, Gumbel-Softmax \cite{jang,maddison}, Semantic Hashing \cite{Salakhutdinov}, Vector quantization \cite{huang},  Improved Semantic Hashing \cite{kaiser}, etc. This section introduces aforementioned discretization techniques used to train discrete autoencoders.

\subsection*{Gumbel-Max Trick}
A random variable ‘$g$’ is purportedly a conventional Gumbel distribution if $g = -\log (- \log (A))$ with $A \sim Unif[0,1]$ \cite{maddison}. The importance of this fact is that any discrete distribution can be parametrized in terms of Gumbel random variables if $ x = \mbox {argmax}(g_i + \log{\pi_i})$, where '$g$' is the random variable and $ (g\_1$...$g\_k) $ are independent and identically distributed samples drawn from standard Gumbel distribution with class probabilities ‘$\pi$’. The Gumbel-Max trick presents an easy and effective technique to produce samples of ‘$z$’ from this categorical distribution in Equ.(7) and Equ. (8) as:

\begin{equation}
z = \mbox {one\_hot} (\mbox {argmax}_i [g_i + \log{\pi_i}]) 
\end{equation}
 
or 

\begin{equation}
z =  \mbox {one\_hot}(x)   
\end{equation}

In other words, it is an algorithm used for drawing samples from a discrete distribution where independent Gumbel perturbations are summed to discrete energy function configurations, thereby returning an $argmax$ configuration of the same energy function. 

The argmax function related to the realizations of discrete distribution and Gumbel samples is not continuous. To overcome this problem, the discrete variables need to be relaxed \cite{maddison}. The required relaxation can be achieved by understanding the fact that all the discrete random variables can be invariably presented as $one-hot$ vectors.

\subsection*{Gumbel-Softmax}

Gumbel Softmax activation function is used as a continuous differentiable approximation to argmax function for generating k-dimensional vectors using a temperature constant \cite{jang,maddison}. During training, this temperature constant is smoothly normalized to approximate samples from a categorical distribution. The parameter gradients of this activation function can be easily computed using the reparameterization trick to make the model differentiable. During the initial phase of training, the variance of gradients is low but biased, and ultimately it becomes high but unbiased.

In view of its application in vector quantized variational autoencoders, this function projects the encoder output to obtain the logits $l$ with discrete code $z_d(x)$ using a learnable projection as given in Equ.(6).

\begin{equation}
z_d(x) = argmax (l_i)  
\end{equation}

Similar to the Gumbel Max, '$g$' is the random variable and $(g\_1$...$g\_k)$ are independent and identically distributed samples drawn from standard Gumbel distribution. The log-softmax of '$l$' is computed in Equ.(7) to obtain sample vectors '$v$'.

\begin{equation}
v= \frac{\exp((l_i + g_i) / \tau)}{\sum_i \exp((l_i + g_i) / \tau)}
\end{equation}

\subsection*{Semantic Hashing}
Semantic Hashing \cite{Salakhutdinov} has become immensely prominent technique that encodes the semantics of a document into a binary vector known as 'hash code'. It represents documents as binary hash codes in such a way that similarity between the documents is directly proportional to the Hamming distance between them. This approach converts the documents to their low dimensional representation comprising series of bits. 

\subsection*{Improved Semantic Hashing}

Improved Semantic Hashing helps in propagating gradients in discrete domain \cite{kaiserbengio}. In this technique, the encoder is flattened using a saturating sigmoid function followed by a simple rounding bottleneck \cite{kaiser}. To discretize the sample vectors, Gaussian noise is added to the sigmoid function for training and during evaluation phase as shown in Equ.(8).

\begin{equation}
     \sigma \sp{\prime} = max(0,min(1, 1.2 \sigma(x)) - 0.1)     
\end{equation}

where $ \sigma \sp{\prime}$ is the saturating sigmoid function before the addition of noise.

When the training begins, the noise $\eta \sim \textit{N}{(0,1)}^D $ is added to the encoder $q_\phi(x)$ on which the sigmoid function is applied to obtain vector $f_e(x)$, refer Equ.(9).

\begin{equation}
f_e(x) = \sigma \sp{\prime}(q_\phi(x) + \eta) 
\end{equation}

The discrete vector representation $ g_e{(x)}_i $ mentioned in Equ.(10) is computed using rounding as:

\begin{equation}
g_e(x)_i =
    \begin{dcases}
        1, & \text {if ${f_e(x)}_i$ > 0.5}  \\
        0, & \text {otherwise}
    \end{dcases}    
\end{equation}

Here, the discrete latent code is analogous to $\tau_{log K}^{-1}(g(x)) $ and the input to the decoder $p_\theta (x)$ is given in Equ.(11) as:
\begin{equation}
p_\theta (x)= e_{h_e(x)}^1 + e_{1- h_e(x)}^2  
\end{equation}
  
which is calculated using two embedding spaces denoted by $e^1$ and $e^2$, and the function $h_e$ is selected randomly to take place for $g_e$ or $f_e$ \cite{kaiserbengio}. 

\subsection*{Vector Quantization} 
We are intrigued to apply vector quantization technique in Variational Autoencoder for the Natural Language Processing (NLP) task as almost all its achievements happen to be outside NLP \cite{oordVQ}.

In vector quantization, a latent embedding space $e=[e_1;...;e_k] \in  R^{K \times D} $  is defined; where $K$ is the size of discrete latent space and $D$ is the dimension of each embedding vector. Input $x$ is passed through an encoder producing output $z_e(x)$. To compute discrete latent variable $z$, $z_e(x)$ is passed through a discretization bottleneck using a nearest neighbour look-up using the shared codebook $E$, as shown in Equ.(12).

\begin{equation}
 q(z=k|x)=
    \begin{dcases}
        1, &\text {for z= $\hat{z}$} \\
        0, &\text {otherwise}
    \end{dcases}   
\end{equation}

where $k= \operatorname*{argmin}_{j \in [K]}{{\parallel {z_e{(x)} - e_j}\parallel}_2}$

 The corresponding vector $e_k$ from the embedding space becomes the input to the decoder, refer Equ.(13).
 \begin{equation}
  z_q^i(x) = e_{k_i}^i   
 \end{equation}

The process of Vector Quantization is highly comparable to KNN algorithm where K-dimensional vectors are mapped to an array of vectors in a ‘codebook’ \cite{vq}. The distance between the sample vector and codebook vector is reduced by maximizing the ELBO and is inversely proportional to their semantic similarity. The training objective function for the same is given in Equ.(14).

\begin{equation}
 L= \log p(x|\hat{z}) + \parallel \mbox sg [z_e(x)] - e_{\hat{z}} \parallel_2^2 
   + \beta \parallel z_e(x) - \mbox sg [ e_{\hat{z}}] \parallel_2^2
\end{equation}

In the above equation, the first term is reconstruction loss which optimizes the encoder and decoder. Second term is VQ or Codebook Loss or L2 error between the embedding space and encoder outputs. Considering the fact that gradients sidestep the embeddings in the backward pass, L2 error is used by a dictionary learning algorithm to transfer the embedding vectors towards the encoder output. The third term is the commitment loss which is a measure to ensure that the encoder output stays close to the embedding space. This is due to the fact that embedding space, whose volume is boundless, can intensify erratically if the embeddings and encoder parameters do not train on the same pace. In this equation, $sg(.)$ stands for stop gradient operator, defined in Equ.(15).

\begin{equation}
\mbox{sg(a)} = 
    \begin{dcases}
    a, & \text{forward computation}\\
    0, & \text{backward computation}
    \end{dcases}     
\end{equation}
  
Here, $\log p(x)$ can be bound by maximizing the ELBO. By defining a simple uniform prior over z, a constant value of KL divergence is obtained given as $\log K$. 

\section{Datasets}
For conducting experiments for multi-class text classification, we have used 3 standard datasets commonly used nowadays for this purpose: AG News \cite{zhang}, TREC6 \cite{voorhees}, and DBPedia \cite{zhang}.

\section{Literature Survey}

Recently, ample research has been carried out in the field of discrete representation learning. Although it has proved successful in speech and vision, not much has been considered in NLP except a translation model by \cite{kaiser} where vector quantization is used to encode an input sequence into discrete code. However, their work targets non-autoregressive decoding to make quick conclusions.

The VQ-VAEs are the building blocks of the research work done in the present paper \cite{oordVQ}. We have analyzed some generic text representations inspired by VQ-VAEs in discrete domain that deal with problems like posterior collapse and index collapse. Although extensive research has been carried out for training VAEs using discrete latent variables, the present section explores some of the relevant works from this huge body of literature.

\subsection*{Conventional VAE}

The VAEs are robust deep generative models extensively used for semi-supervised learning \cite{kingma,rezende}. They model the data distribution and implictly understand the likeness between the instances. Originally introduced by \cite{kingmaVAE,rezende} in 2013, the VAEs capture the relevant features for training continuous latent variables by optimizing the ELBO. For instance, Bowman et al. \cite{bowman} trained a novel VAE which refrained the KL divergence term to minimize by annealing it in ELBO from 0 to 1 gradually over time. This helped to avoid the issue of posterior collapse to some extent. They also introduced an RNN model based on VAE that grasps global features from the text and reconstructs them by decoding efficaciously. Lately, TVAE model \cite{tvae} was introduced for performing sentiment analysis using VAE. Considering this fact, they proposed a modification in VAE to minimize the inference lag in order to optimize the inference network. This work was extended by Takida et al. \cite{takida}. According to the authors, diminished KL divergence is not the sole cause of posterior collapse. This cause generally relates to hyperparameters resembling the data variance whose irrelevant choice causes oversmoothness, leading to posterior collapse. Hence, they presented AR-ELBO which controls the smoothness of the model by adjusting the variance parameter.

\subsection*{Discrete Latent Variables}

Recently, there has been a lot of research for training discrete latent variable models. Out of this, a great deal mainly focuses on limiting the ELBO using sampling-based techniques \cite{mnih}. However, utilizing discrete latent variables is a challenging task as even now most of the work is dominated by continuous latent variables. To meet this challenge, a few authors proposed a new continuous reparameterization trick based on Gumbel Softmax distribution \cite{jang}. This trick is basically a continuous distribution having a temperature constant annealed amid training that converges to a discrete distribution in the limit. However, it does not bridge the performance gap of VAEs with continuous latent variables where Gaussian reparameterization trick can be used. This is so because this trick benefits from much lower variance in the gradients and is typically evaluated on smaller databases with low dimensionalities (below 7). 

Many alternative strategies have been devised for training discrete VAEs. For instance, the NVIL estimator introduced by Andriy et al. \cite{mnih} optimizes the ELBO using a single sample objective. This estimator uses several techniques to reduce variance and accelerate the training process. On the other hand, \cite{mnih2} introduced “VIMCO” which uses a multi-sample objective from \cite{burda}. This objective further accelerates the convergence by using numerous samples from the inference network.

Another promising technique used by researchers for discrete latent variable models is to use diverse gradient estimators. Initially, Williams et al. \cite{williams} designed an  unbiased and high variance REINFORCE estimator. Apart from the gradient estimators, an alternate approach is Gumbel-Max trick \cite{gumbelmax}. As mentioned in Subsection C. of Section 2, the argmax function related to the realizations of discrete distribution and Gumbel samples is not continuous. To overcome this, the discrete variables need to to be relaxed. As a result, \cite{jang,maddison} introduced  Gumbel-Softmax reparameterization trick to use continuous moderation of categorical distributions, like Gumbel-Softmax reparameterization trick. This trick provides biased gradients with low variance. In 2018, Lukasz et al. \cite{kaiserbengio} proposed an improved version of semantic hashing known as Improved Semantic Hashing. This discretization technique helps to propagate gradients in discrete domain\cite{kaiser}. 

Lin et al. \cite{lin} proposed to use Boltzmann-machine (BM) distribution as variational posterior to establish correlations among bits of hashcodes. The promising generative semantic hashing technique introduced by the authors addresses the intractability issue of training by augmenting the BM distribution as a hierarchical concatenation of Gaussian-like and Bernoulli distribution.

\subsection*{Vector Quantization}

The Variational Autoencoder (VAE) based on Vector Quantization (VQ) technique was proposed by Aaron et al. \cite{oordVQ}, thus called as VQ-VAE. Recently, a non-autoregressive model was trained to accelerate the decoding step. The authors used a self-attention based Transformer model \cite{vaswani} along with REINFORCE algorithm \cite{williams} to address the multi-modality problem for modeling the ‘fertilities’ of words in the sequences. However, comprehensive fine-tuning is required for policy gradients to work in REINFORCE and being a generic approach, it can not be applied to other sequence learning tasks. Based on the vector quantization, \cite{shen}  proposed a prototype based concept to deal with the problems of Online Semi-Supervised Learning (OSSL). For improving these prototypes in OSSL and boosting their performance, two efficient strategies were used, namely maximum conditional likelihood method for labeled data and Gaussian mixture for unlabeled data. Since Information Bottleneck (IB), a special class of vector quantization which can benefit from this technique by learning the representations of multiple inputs concurrently and when reinforced with variational strategies makes way for ‘Aggregated Learning’ framework \cite{sofl}.

\section{Proposed Model}

\begin{figure}[!htb]
%\minipage{0.50\textwidth}
  \includegraphics[width=\linewidth]{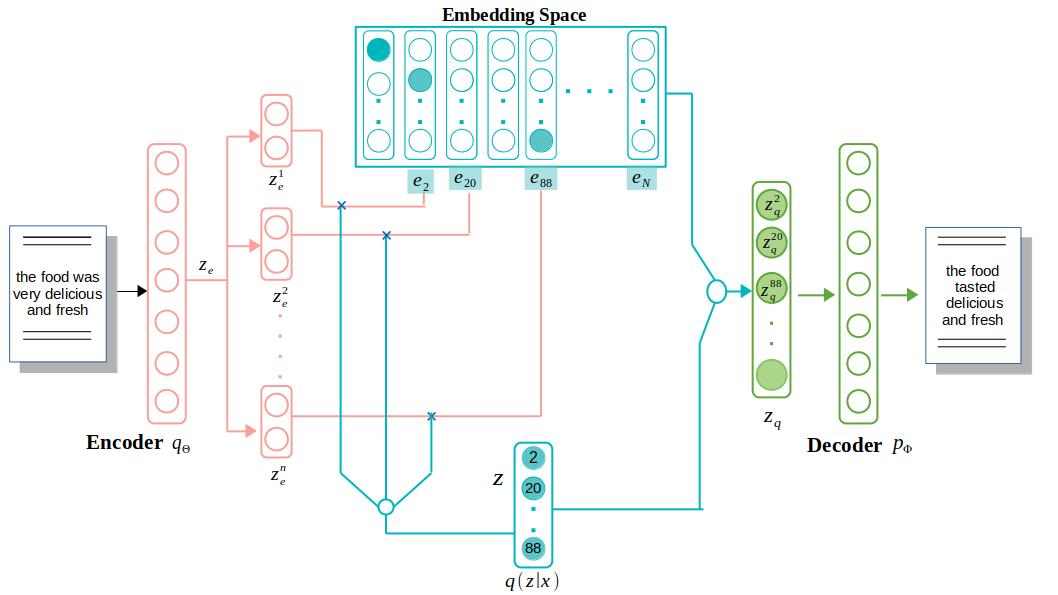}
  \caption{Proposed model architecture illustrating an encoder-decoder setup based on decomposed vector quantization. The encoder is decomposed into '$n$' sub-encoders whose output is mapped to a nearest point in the Embedding Space. The corresponding indices are sent to decoder for reconstruction. The gradients of Loss function are adaptively modified using VQ Loss for backpropagation. Refer Fig. ~\ref{backpropagation} for insight view of backpropagation.}\label{proposed_model}
%\endminipage
%\hfill

\end{figure}

In this section, we are going to present the framework of our model regarding its structure and execution. The structure of our model marks an advanced way for training VAEs \cite{kingma,rezende} with discrete latent variables for the multiclass text classification task and deals with a challenging and long-standing issue like latent variable collapse (or posterior collapse). The proposed model also addresses the index collapse issue efficiently by modifying the training scheme so that the latent variables are not ignored. The complete framework of our model is shown in Figure ~\ref{proposed_model}.

During forward pass, an input sequence ‘$x$’ which is passed to the encoder. The encoder uses Transformer layer network \cite{vaswani} to map the data instance  $x$ to parameters of an approximate posterior distribution and produces output $z_e(x)$. Using the decomposed vector quantization technique, the encoded output is split into '$n$' sub-encoders: $z_e^1(x) \circ z_e^2(x) ... \circ z_e^n(x)$, where '$\circ$' indicates concatenation operator.

A latent embedding space, also known as codebook $e \in R^{K \times D}$  is defined, where $D$ is the dimension of each latent embedding vector $e_i$, and K is the size of embedding space. For each $z_e^i(x)$, embedding space $e^i \in R^{K\sp{\prime} \times {D/n}} $ is alloted where $K\sp{\prime} = 2^{{\log_2 K}/n} $. The posterior distribution probabilities $q(z|x)$ is defined in Equ. (16) as:

\begin{equation}
 q(z|x)=
\begin{dcases}
    1, &\text{for z= $\hat{z}$}  \\
    0, &\text{otherwise}
\end{dcases}   
\end{equation}

where $\hat {z}_i= \operatorname*{argmin}_{j \in [K]}{{\parallel {z_e^i{(x)} - e_j^i}\parallel}_2}$

The output of sub-encoders $z_e^i$ is passed through a discretization bottleneck and mapped onto nearest codebook embeddings. Manhattan distance, also known as L1 metric, is used to calculate the mapping distance, and is given by Equ. (17).

\begin{equation}
   \mbox{Manhattan Distance} = |a_1 - a_2| + |b_1 - b_2|
\end{equation}

where data point $d_1$ is at coordinates $(a_1, b_1)$ and its nearest neighbour $d_2$ at coordinates $(a_2, b_2)$.

For the above mentioned codebook embeddings, Shannon Entropy is calculated as shown in Equ. (18).
\begin{equation}
    e(z_e^i) = - \sum_{c \in C} p_c(z_e^i)*log(p_c(z_e^i))
\end{equation}
where $e(z_e^i)$ is the Shannon entropy of each sub-encoder and $p_c(z_e^i)$ is the corresponding probability for all the possible classes denoted by $C$ in Equ. (19).

DropConnect technique is applied to the sub-encoders having entropy higher than a certain experimental threshold value. This threshold entropy is the median of all entropy values of each sub-encoder. This technique proves to be effective to preserve maximum information by applying DropConnect. The resulting embeddings are concatenated to calculate the discrete latent variable $z_q(x)$ as $z_q^1(x) \circ z_q^2(x) ... \circ z_q^n(x)$, where '$\circ$' represents the concatenation operator over '$i$' instances.

\begin{figure}
%   \minipage{0.50\textwidth}
  \includegraphics[width=0.75\linewidth, height=0.15\textheight]{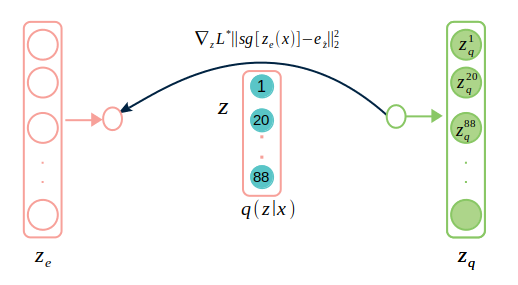}
  \caption{Illustration of backpropagation method in the proposed model. Refer Fig. ~\ref{proposed_model} for details of components.}\label{backpropagation}
%\endminipage\hfill
\end{figure}

The indices of the codebook embeddings become input for the decoder which gives a reconstructed input as output. In the backward pass, the gradients of Loss function $\nabla _z L$ are calculated with respect to the model parameters. These gradients are then adaptively modified using $\parallel \mbox sg [z_e(x)] - e_{\hat{z}} \parallel_2^2 $, which is the 'VQ Loss' calculated in the forward pass, refer Fig. \ref{backpropagation}. We use this modification in place of passing unaltered gradients in the backward pass to preserve information of the fine-tuned weights to minimize the loss. This information is fedback to the input to know the effects of each node towards the loss and make the model reliable. Gradient Clipping technique along with a predetermined gradient threshold is used to avoid the problem of exploding gradients.

For the classification task, the pretrained encoder is freezed and a classifier is added on top of it. This classifier is utilized for probability estimation over the label distribution. It uses task-specific parameters and variable amount of labeled instances (5\%, 20\%, 50\% and 100\%) for all the databases.

Additionally, the present paper also incorporates a class of VQ-VAE that amends the embedding space with Exponential Moving Averages (EMA). This class is independent of the choice of optimizer to learn the embedding and converges swiftly. Other than vector quantization, EMA has proved to be an efficient method for learning the prior distribution without using the gradient descent technique. The training objective function in Equ. (7) remains the same, besides two quantities:

\begin{itemize}
    
    \item Number of hidden states $c_j$ of encoder with $e_j$ number of nearest neighbour are shown in Equ. (19). The count of these hidden states is updated over variable amount of labeled instances.
    
    \item Individual embedding spaces $e_i$ being trained using EMA from $z_e^i(x)$ over variable amount of labeled instances as given in Equ. (20).
\end{itemize}

\begin{equation}
    c_j^i \leftarrow \lambda c_j^i + (1-\lambda) \sum_{l} \delta [z_q(x_l) = e_j^i]
\end{equation}

\begin{equation}
    e_j^i \leftarrow \lambda e_j^i + (1-\lambda) \sum_{l} {\frac{\delta [z_q^i(x_l) = e_j^i] z_e^i(x_l)}{c_j^i}}
\end{equation}

where $\lambda$ is the decay parameter and $\delta$[.] the indicator function.

\section{Experimental Details}

\subsection{Hyperparameters}

The proposed model is implemented using PyTorch package. For preprocessing the text, we used spaCy from AllenNLP. Adam optimizer has been employed for all the experiments with a 1e-4 learning rate, 1e-9 epsilon, and 0.9 and 0.98 $\beta1$ and $\beta2$ respectively. For model regularization, DropConnect with 0.1 rate has been applied to word embeddings. While using exponential moving averages in our model, we have experimented using a decay parameter with value 0.99 and epsilon 1e-9. Kaiming uniform \cite{kaiming} has been employed for initializing the codebook vectors. 
To evaluate the efficiency of the latent representation, a classifier has been trained with varying numbers of labeled instances: 5\%, 20\%, 50\% and the 100\% (varies by dataset). We have used accuracy, precision, recall and micro F-score as the evaluation metrics.

\section{Experimental Results}

The present section comprises of experimental results of our model for DBPedia database. Results of both Regular and EMA versions of the model have been incorporated. We have computed the classification results for variable amount of labeled instances. Table ~\ref{db_results} illustrates VQ Loss, Accuracy, Precision, Recall and F1 score in detail for DBPedia dataset.

\begin{table}[!t]
\renewcommand{\arraystretch}{0.8}

  \caption{Classification results of both Regular and EMA version of our proposed model when computed for variable amount of labeled instances for DBPedia dataset} 
    \label{db_results} 
    
     \centering
    \resizebox{0.8\linewidth}{!}{
    \begin{tabular}{c c c c c c c c}
    \hline
    Dataset & \shortstack{Type of \\model} & \shortstack{Number of \\samples} & Loss & Accuracy & Precision & Recall & F1 Score \\
    \hline
    
        & 
            & 5\% & \shortstack{0.196} & \shortstack{94.35} & \shortstack{93.82} & \shortstack{93.22} & \shortstack{93.51}\\ 
        & 
            & 20\% & \shortstack{0.109} & \shortstack{97.09} & \shortstack{97.12} & \shortstack{97.23} & \shortstack{97.18} \\ 
        &    Regular
            & 50\% & \shortstack{0.102} & \shortstack{97.42} & \shortstack{97.35} & \shortstack{97.33} & \shortstack{97.34} \\ 
          \multirow[b]{2}{*}{DBPedia}   
            & & 100\% & \shortstack{0.055} & \shortstack{98.67} & \shortstack{98.65} & \shortstack{98.65} & \shortstack{98.65} \\   
            \cline{2-8}

        & 
            & 5\% & \shortstack{0.19} & \shortstack{96.08} & \shortstack{96.34} & \shortstack{96.25} & \shortstack{96.29}\\ %\noalign{\smallskip}
        &
            & 20\% & \shortstack{0.11} & \shortstack{97.16} & \shortstack{97.42} & \shortstack{97.34} & \shortstack{97.38} \\
        &  EMA  
            & 50\% & \shortstack{0.104} & \shortstack{97.55} & \shortstack{97.56} & \shortstack{97.56} & \shortstack{97.56}\\
        &    
            & 100\% & \shortstack{0.042} & \shortstack{99.12} & \shortstack{98.92} & \shortstack{98.92} & \shortstack{98.92}\\
    
    \hline
    \end{tabular}
  }
\end{table}

%\clearpage
\subsection{Comprehensive comparison}

In this section, we present the classification results of our proposed model with the existing state-of-the-art models. This comparison has been illustrated in Table ~\ref{table_comparison}. Several versatile deep and non-deep learning models have been selected with same criteria for the comparison. This means that the classification results have been calculated for variable number of labeled instances (5\%, 20\%, 50\% and 100\%) and on similar hyper-parameters for all the models mentioned in Table ~\ref{table_comparison}. Highest accuracy values have been highlighted in bold for a better view.

\begin{table}[!t]
\renewcommand{\arraystretch}{0.85}

\caption{Performance comparison of our proposed model (both versions) to existing state-of-the-art models on the basis of classification accuracy for DBPedia dataset. Highest accuracy values are highlighted in \textbf{bold}}

    \label{table_comparison}
    
    \centering
        \resizebox{0.75\linewidth}{!}{
      \begin{tabular}{c c c c c c}
    \hline%\noalign{\smallskip}
    \multirow{2}{*}{Dataset} & \multirow{2}{*}{Model} &
      \multicolumn{4}{c|}{Variable number of labeled instances}\\
      \cline{3-6}
       
    & & 5\% & 20\% & 50\% & 100\%   \\  
    \hline

    \multirow{10}{*}{DBPedia} & BiLSTM \cite{lstm} & 92.27\% & 94.29\% & 93.69\% & 95.32\% \\
    
    & GloVe \cite{glove} & 93.11\% & 94.45\% & 96.72\% & 97\%  \\
    
    & CBOW \cite{cbow} & 94.67\% & 94.87\% & 95.16\% & 96.63\%\\
    
    & Local \cite{shun} & 93.43\% & 95.61\% & 95.98\% & 97.12\%\\
    
    & VAMPIRE \cite{vampire} & 95.25\% & 96.88\% & 97.25\% & 98.20\%\\
    
    & Transformer \cite{vaswani} & 94.34\% & 93.52\% & 93.87\% & 96.91\%\\
    
    & fastText \cite{fasttext} & 93.23\% & 94.79\% & 96.68\% & 97.83\%\\
    
    & CluE-CVAE \cite{clue} & 92.98\% & 93.45\% & 94.55\% & 98.71\%\\
    
    & {Our model (Regular)} & {94.35}\% & {97.09}\% & {94.72}\% & \textbf {98.65}\%\\
    
    & {Our model (EMA)} & {96.08}\% & {97.16}\% & {97.66}\% & \textbf {99.12}\%\\
    \hline
  \end{tabular}
  }
  %\end{center}
\end{table}

From Table ~\ref{table_comparison}, it can be observed that performance of both Regular and EMA versions of our proposed model has been compared to both Machine and Deep Learning models. To commence with some machine learning models such as BiLSTM, GloVe, CBOW, fastText, etc for the DBPedia dataset, it can be observed that large amount of data has been required by all the models to obtain superior results. Here too the deep learning models have performed well as compared to their machine learning equivalents as they have an added advantage of pretraining. The proposed varients are trained on complete training data to compete with existing state-of-the-art models and outclass them.

\section{Conclusion}
In this manuscript, we have proposed a model inspired from vector quantization technique which focuses on semi-supervised multi-class text classification and generation task. Both experimental and analytical work has been carried out to address problems like Posterior collapse and Index collapse through proposed strategies. Transformer based encoder-decoder architecture has been utilized as the backbone of our proposed model. 

Moreover, the loss gradients are not passed to the encoder unaltered as followed in original research \cite{oordVQ} during backpropagation. Rather they are adaptively modified by VQ Loss, thereby enhancing the performance of our model by transferring the gradients with additional information. Three experimental datasets have been selected to validate the proposed model and it has been proved that our model has achieved remarkable results when compared to the existing baselines. 

In future, we plan to continue our research in the same direction to explore other complex semi-supervised NLP tasks.

\section*{Acknowledgments}
This work was supported by free academic credits from Google Cloud Platform.

%Bibliography
\bibliographystyle{unsrt}  
\bibliography{references}

\end{document}